\documentclass[10pt,twocolumn,letterpaper]{article}

\usepackage{iccv}
\usepackage{times}
\usepackage{epsfig}
\usepackage{graphicx}
\usepackage{amsmath}
\usepackage{amssymb}

\usepackage{multirow}
\usepackage{subfigure}
\usepackage{algorithm}
\usepackage{algorithmic}

\usepackage[breaklinks=true,bookmarks=false]{hyperref}

\iccvfinalcopy 

\def\iccvPaperID{****} 

\begin{document}

\title{Learning Mid-level Words on Riemannian Manifold for Action Recognition}

\author{Mengyi Liu, Ruiping Wang, Shiguang Shan,  Xilin Chen\\
Key Lab of Intelligent Information Processing of Chinese Academy of Sciences (CAS),\\
Institute of Computing Technology, CAS, Beijing, 100190, China\\
{\tt\small mengyi.liu@vipl.ict.ac.cn, \{wangruiping, sgshan, xlchen\}@ict.ac.cn}
}

\maketitle

\vspace{-10pt}
\begin{abstract}
Human action recognition remains a challenging task due to the various sources of video data and large intra-class variations. It thus becomes one of the key issues in recent research to explore effective and robust representation to handle such challenges. In this paper, we propose a novel representation approach by constructing mid-level words in videos and encoding them on Riemannian manifold. Specifically, we first conduct a global alignment on the densely extracted low-level features to build a bank of corresponding feature groups, each of which can be statistically modeled as a mid-level word lying on some specific Riemannian manifold. Based on these mid-level words, we construct intrinsic Riemannian codebooks by employing K-Karcher-means clustering and Riemannian Gaussian Mixture Model, and consequently extend the Riemannian manifold version of three well studied encoding methods in Euclidean space, i.e. Bag of Visual Words (BoVW), Vector of Locally Aggregated Descriptors (VLAD), and Fisher Vector (FV), to obtain the final action video representations. Our method is evaluated in two tasks on four popular realistic datasets: action recognition on YouTube, UCF50, HMDB51 databases, and action similarity labeling on ASLAN database. In all cases, the reported results achieve very competitive performance with those most recent state-of-the-art works.
\end{abstract}

\vspace{-10pt}
\section{Introduction}

Recognizing human actions in videos has been a popular research field in recent years with a wide range of applications, including video surveillance, human-computer interaction, etc. Recent research mainly focuses on the realistic datasets collected from web videos or digital movies \cite{kliper2012action,kuehne2011hmdb,liu2009recognizing}. These real-world scenarios impose great challenges for action recognition, e.g. the high-dimension of video data and large intra-class variations caused by scale, viewpoints, illumination.

To deal with such challenges, many researchers attempt to explore effective and robust video representation methods for dynamics modeling. One popular family of approaches employ low-level representation schemes, e.g. local space-time descriptors (STIP\cite{laptev2005space}, 3D SIFT \cite{scovanner20073}, Extended SURF \cite{willems2008efficient}, HOG3D \cite{klaser2008spatio}), dynamic texture (LTP \cite{yeffet2009local}, MIP \cite{kliper2012motion,hanani2013evaluating}), and optical flow based method (MBH \cite{dalal2006human}, Dense Trajectory \cite{wang2013dense,wang2013action}). A bag-of-features encoding of these low-level features can be directly used for action recognition and some of them have reported the state-of-the-art performance \cite{wang2013dense,wang2013action}. Another line of research extracts high-level information of human motion shape by constructing explicit models of bodies, silhouettes, or space-time volumes \cite{bobick2001recognition,cheung2003shape,gorelick2007actions}. More recently, Action Bank \cite{sadanand2012action}, is proposed to construct a bank of action templates for high-level representation, which leads to good performance and also possesses semantic meaning.

While certain success has been achieved in both two lines mentioned above, there still remains several unresolved limitations. For low-level features, the local patterns are repeatable and thus robust to intra-class variation, but lack of descriptive and discriminative ability. In contrast, high-level features possess the global semantic information, which however simultaneously brings sensitivities to unprofitable variations and deformations. To balance between the low-level and high-level, a couple of recent works propose to learn mid-level representations, e.g. Action-Gons \cite{wang2014action}, Actons \cite{zhu2013action}, motionlets \cite{wang2013motionlets}, motion atoms \& phrases \cite{wang2013mining}, which are expected to possess both local repeatability and global descriptive ability. In both \cite{wang2013motionlets} and \cite{wang2013mining}, the mid-level units are constructed by clustering groups of low-level features, then the video representation is obtained by encoding these mid-level features using activation or correlation functions.

\begin{figure*}[tbh]
\centering
\includegraphics[height=7.5cm]{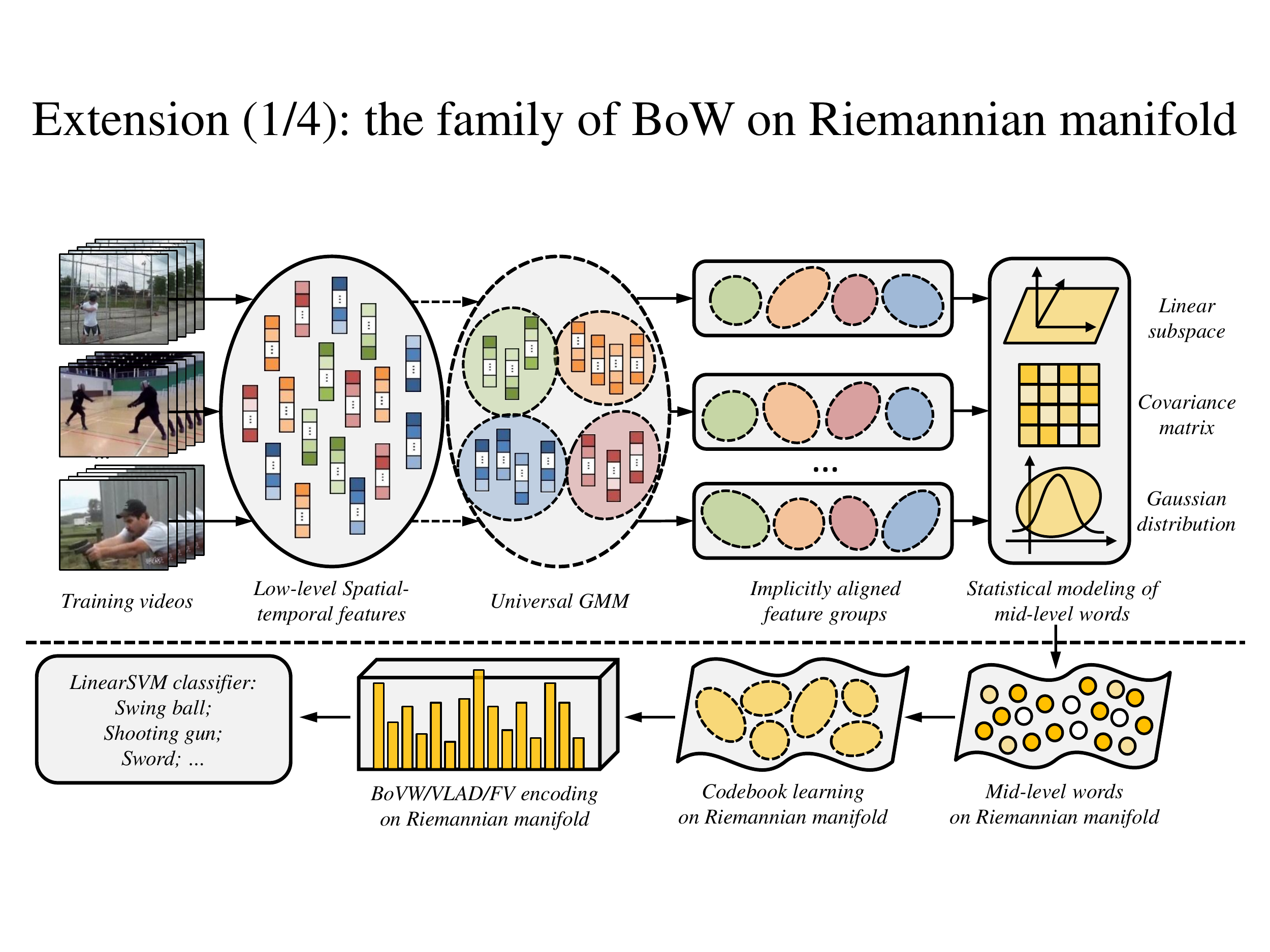}
\caption{An overview of the proposed method. we first conduct a global alignment on the densely extracted low-level features via universal GMM to construct a bank of corresponding feature groups. For each implicitly aligned feature group (corresponding to certain Gaussian component), we employ three types of statistics, i.e. linear subspace, covariance matrix, and Gaussian distribution, to statistically model it as a mid-level word lying on Riemannian manifold. Then by construction of intrinsic Riemannian codebooks, three encoding methods BoVW/VLAD/FV can be extended to Riemannian manifold respectively for the final representations. \textbf{Best viewed in color}.}
\label{figure:figoverview}
\end{figure*}

In light of such progresses, in this paper, we propose a novel mid-level representation with several encoding methods to further improve the performance. Compared with the former work, our contribution lies in three aspects: (1) We consider a global alignment among video samples to build semantic correspondence for matching and alleviate the influence of unexpected noise caused by realistic scenario; (2) We employ three types of statistics from different perspectives, i.e. linear subspace, covariance matrix, and Gaussian distribution, for mid-level words modeling, which introduce some non-Euclidean spaces, i.e. Riemannian manifolds, for feature diversity; (3) Different encoding methods on Riemannian manifold are investigated for effective video representation and further improving the performance. An overview of our method is illustrated in Figure~\ref{figure:figoverview}. The upper part shows the procedure of mid-level words construction: groups of low-level features are globally aligned via GMM and then modeled as mid-level words statistically. The lower part depicts the mid-level words encoding on Riemannian manifold: intrinsic Riemannian codebooks are generated and different encoding methods can be conducted on Riemannian manifold for the overall video representations. Finally, linear SVM is employed for recognition.

Our method is evaluated in two tasks on four realistic datasets: action recognition on YouTube \cite{liu2009recognizing}, UCF50 \cite{reddy2013recognizing} and HMDB51 \cite{kuehne2011hmdb} databases, and action similarity labeling on ASLAN database \cite{kliper2012action}. The experiments demonstrate the benefits of encoding mid-level words on Riemannian manifold compared with the original descriptors. By fusing descriptors and different mid-level words modeling strategies, our results achieve the state-of-the-art on all databases.

\section{Mid-level words construction}

\subsection{Low-level features}

To construct the mid-level words, we first need to extract the low-level spatial-temporal features. As dense features have shown to improve the classification performance in a few recent works \cite{wang2009evaluation,wang2013dense}, in this work, we sample the feature points on dense space-time grid, and exploit several state-of-the-art descriptors: Histograms of Oriented Gradient (HOG) \cite{dalal2005histograms}, Histograms of Flow (HOF) \cite{laptev2008learning}, and Motion Boundary Histograms (MBH) \cite{dalal2006human}. These descriptors capture the low-level appearance/dynamic information from different views and are expected to complement with each other. HOG mainly focuses on static appearance, while HOF captures the local motions. Besides the motion objects, these two descriptors can effectively encode the spatial context information for some environment-related actions, like sports \cite{wang2013dense}. The MBH is obtained by computing derivatives of the optical flow, which encodes the relative motion between pixels, and thus provides a simple way to discount for camera motions.

\subsection{Global alignment}

For each type of low-level feature above, one video sample can be represented as a local feature set. To handle the large intra-class variations, a video alignment scheme is definitely required to build semantic correspondence for matching and alleviate the influence of unexpected noise caused by realistic scenario. Inspired by \cite{hasan2011study}, we employ Gaussian Mixture Model (GMM) on all of the local features to learn a global alignment model which can statistically unify all the local variations presented in the video samples, and thus facilitate a robust parameterized modeling of each feature set (i.e. video). Here we simply build a GMM with spherical Gaussian components as
\begin{equation}
P(f|\Theta)=\displaystyle{\sum_{k=1}^K}w_k G(f|\mu_k,\sigma_k^2I),
\end{equation}
where $\Theta=(w_1,\mu_1,\sigma_1,...,w_K,\mu_K,\sigma_K)$ and $f$ denotes the low-level feature; $K$ is the number of Gaussian mixture components; $I$ is identity matrix; ${w_k,\mu_k,\sigma_k^2}$ are the mixture weight, mean, and diagonal covariance of the $k$-th Gaussian component $G(f|\mu_k,\sigma_k^2I)$.

We use classical Expectation-Maximization (EM) algorithm to estimate the parameters by maximizing the likelihood of the training feature set. After building the GMM, each video can also be represented as $K$ corresponding feature groups by fitting its feature set to each universal Gaussian component and thus be aligned manually. Formally, suppose we have $N$ videos, the $i$-th video sample $V^i(i=1,2,...,N)$ can be represented as a local feature set $\mathcal{F}^i=\{f^i_1,f^i_2,...,f^i_{L_i}\}$, where $f^i_l$ belongs to $R^d$ is a $d$-dimensional feature vector; $L_i$ is the number of low-level features in $V^i$. For the $k$-th Gaussian component, we can calculate the probabilities of all $f^i_l$ in $\mathcal{F}^i$ as
\begin{equation}
\begin{array}{rl}
P_k^i=\{p_k(f^i_l)~|~p_k(f^i_l)=w_kG(f^i_l|\mu_k,\sigma_k^2I)\}_{l=1}^{L_i}.\\
\end{array}
\end{equation}

By sorting the elements of $P^i_k$ in descending order, the features with the largest $T$ probabilities are selected out to construct a compact feature group, which can be represented as $F^i_k=\{f_{k_1}^i,...,f_{k_T}^i\}$. Therefore, by fitting to the same Gaussian component, the feature groups $\{F^1_k,F^2_k,...,F^N_k\}$ for all $N$ videos are implicitly aligned with appearance or semantic correspondence.

\subsection{Mid-level words modeling}

According to the construction scheme of each $F^i_k$, the local features $\{f_{k_1}^i,...,f_{k_T}^i\}$ in the group are expected to share similar appearance and close space-time location. From a view of statistical modeling, the dynamic information within the feature group can be favorably characterized by exploring correlations and variations among those low-level features. In this study, each feature group is modeled as a space-time pattern, namely mid-level word, using three statistics from different aspects: linear subspace \cite{hamm2008grassmann}, covariance matrix \cite{tuzel2008pedestrian,wang2012covariance}, and Gaussian distribution \cite{arandjelovic2005face}.

\textbf{Linear subspace}. By assuming data samples share a certain degree of linear correlations, the feature set $F^i_k=\{f_{k_1}^i,...,f_{k_T}^i\}$ can be represented by a low-dimensional linear subspace $U^i_k\in R^{d\times r}$ via SVD as follows:
\begin{equation}
\sum_{t=1}^T (f_{k_t}^i-\overline{f_k^i})(f_{k_t}^i-\overline{f_k^i})^T=U^i_k \Lambda^i_k {U^i_k}^T,
\end{equation}
where $\overline{f_k^i}$ is the mean of the local feature vectors in $F^i_k$. $U^i_k=[u_1, u_2,...,u_r]$, $u_j$ is the $j$-th leading eigenvector, and $r$ is the dimension of the subspace. Similar idea of using ``subspace'' also appears in the Local Coordinate System \cite{delhumeau2013revisiting} for VLAD, where ``subspace'' served as a mapping for residual vector, while our method directly takes ``subspace'' for further encoding to retain the entire data structure.

\textbf{Covariance matrix}. We can also represent the feature set with the $d\times d$ sample covariance matrix:
\begin{equation}
C_k^i=\frac{1}{T-1}\displaystyle{\sum_{t=1}^T}(f_{k_t}^i-\overline{f_k^i})(f_{k_t}^i-\overline{f_k^i})^T,
\end{equation}
The main difference between $U^i_k$ and $C^i_k$ is whether discarding the eigenvalues or not during SVD. Eigenvalues capture the relative importance (magnitude) of different variance directions. It is also well known that the $d\times d$ nonsingular covariance matrices are Symmetric Positive Definite (SPD) matrices $Sym_d^+$ lying on a Riemannian manifold \cite{arsigny2007geometric}.

\textbf{Gaussian distribution}. Suppose the feature vectors $\{f_{k_1}^i,...,f_{k_T}^i\}$ follow a $d$-dimensional Gaussian distribution $\mathcal{N}^i_k(\mu,\Sigma)$, where $\mu$ and $\Sigma$ are the data sample mean ($\mu=\overline{f_k^i}$) and covariance ($\Sigma=C_k^i$) respectively. According to information geometry, the space of $d$-dimensional multivariate Gaussians is also a specific Riemannian manifold and can be embedded into the space of Symmetric Positive Definite (SPD) matrices Riemannian manifold\cite{lovric2000multivariate}, denoted as $Sym_{d+1}^+$. In particular, a $d$-dimensional Gaussian $\mathcal{N}^i_k(\mu,\Sigma)$ can be uniquely represented by a $(d+1)\times(d+1)$ SPD matrix $G^i_k$ as follows:
\begin{equation}
\mathcal{N}^i_k(\mu,\Sigma)\sim G^i_k = |\Sigma|^{-\frac{1}{d+1}}\begin{bmatrix}
\Sigma+\mu\mu^T & \mu \\
\mu^T & 1
\end{bmatrix}
\end{equation}
Thus we can measure the intrinsic geodesic distance between Gaussians on the underlying Riemannian manifold in the same way as that between SPD matrices.

\section{Mid-level words encoding}

As encoding methods (e.g. Bag of Words, Fisher Vectors) applying on low-level features are shown to be effective for action recognition \cite{wang2013action,oneata2013action,cai2014multi,peng2014action,peng2014large}, in this work, we extend the Euclidean encoding methods to Riemannian manifold for mid-level words based representation.

\subsection{Riemannian codebook}

To encode the three types of mid-level words, which all reside on some certain Riemannian manifolds, we first need to construct the Riemannian codebook. To this end, K-means and GMM are two well studied and widely used techniques in vector space. To make them applicable to Riemannian manifold, in this section, we derive their extensions as K-Karcher-means and Riemannian GMM. For ease of presentation, we denote different mid-level words $U^i_k$, $C^i_k$, and $G^i_k$ all by $X^i_k$.

\textbf{K-Karcher-means}. Traditional K-means partitions the space of feature vectors into informative regions based on Euclidean metrics. Considering the geometrical structure of $X^i_k$, if one simply computes their Euclidean sample means as doing for general matrices in Euclidean space, such means will obviously do not preserve the property of orthonormality or symmetric positive definiteness. Following the recent works studying on Riemannian clustering \cite{turaga2011statistical,faraki2014fisher,tabia2014covariance}, we employ the Karcher mean \cite{karcher1977riemannian} on Riemannian manifold for our purpose. Formally, given a set of mid-level words $X^i_k$, the Karcher mean is defined as the point on the manifold that minimizes the sum of squared geodesic distances:
\begin{equation}
\widehat{X}=\arg \min_{X\in \mathcal{M}}\sum_{i,k} d_g^2(X^i_k,X),
\end{equation}
where $\mathcal{M}$ denotes the Riemannian manifold, and $d_g: \mathcal{M} \times \mathcal{M} \rightarrow R^+$ is the geodesic distance defined on the manifold. Specifically, $d_g$ can be measured using two operators, namely exponential map $exp_X(\cdot)$ and logarithm map $log_X(\cdot)$, defined to switch between the manifold and its tangent space at $X$. Thus $\widehat{X}$ is the solution to $\sum_{i,k} log_X(X^i_k)=0$, which can be solved iteratively as in Algorithm~\ref{alg:algKarcher}.

\begin{algorithm}[tbh]
\renewcommand{\algorithmicrequire}{\textbf{Input:}}
\renewcommand\algorithmicensure {\textbf{Output:}}
\caption{\textbf{:~Karcher mean}}
\begin{algorithmic}[1]
\label{alg:algKarcher}
\vspace{3pt}
\REQUIRE ~~\\
Mid-level words subset:~$\{X^i_k\}$\\
\vspace{3pt}
\ENSURE ~~\\
Karcher mean:~$\widehat{X}$
\vspace{6pt}
\STATE Set initial estimate of the Karcher mean as $\widehat{X}^{(0)}$ \\
\vspace{3pt}
\STATE Set $p=0$ \\
\vspace{3pt}
\STATE \textbf{while} $p<max\_iter$ \textbf{do}\\
\vspace{3pt}
\STATE ~~~~For each $X^i_k$, compute the tangent vector:\\
\vspace{3pt}
~~~~$v^i_k=log_{\widehat{X}^{(p)}}(X^i_k)$
\vspace{3pt}
\STATE ~~~~Compute the mean vector $\overline{v^i_k}$ in tangent space\\
\vspace{3pt}
\STATE ~~~~\textbf{if} $||\overline{v^i_k}||^2$ is small enough,
\vspace{3pt}
\STATE ~~~~ ~~~~\textbf{break};\\
\vspace{3pt}
\STATE ~~~~\textbf{else}\\
\vspace{3pt}
\STATE ~~~~ ~~~~Move $\overline{v^i_k}$ back to manifold:\\
\vspace{3pt}
~~~~ ~~~~$\widehat{X}^{(p+1)}=exp_{\widehat{X}^{(p)}}(\overline{v^i_k})$\\
\vspace{3pt}
\STATE ~~~~\textbf{end if}\\
\vspace{3pt}
\STATE ~~~~$p=p+1$
\vspace{3pt}
\STATE \textbf{end while}
\end{algorithmic}
\end{algorithm}

To extend K-means to K-Karcher-means, we seek to estimate $M$ clusters with centers $\{\widehat{X}_m\}_{m=1}^M$ such that the sum of squared geodesic distances $\sum_{m=1}^M d_g^2(X^i_k,X_m)$ over all clusters are minimized. The solving procedure is in Algorithm~\ref{alg:algK}.

\begin{algorithm}[tbh]
\renewcommand{\algorithmicrequire}{\textbf{Input:}}
\renewcommand\algorithmicensure {\textbf{Output:}}
\caption{\textbf{:~K-Karcher-means}}
\begin{algorithmic}[1]
\label{alg:algK}
\vspace{3pt}
\REQUIRE ~~\\
Mid-level words training set:~$\{X^i_k\}$\\
\vspace{3pt}
\ENSURE ~~\\
$M$ Karcher cluster centers:~$\{\widehat{X}_1,\widehat{X}_2,...,\widehat{X}_M\}$
\vspace{6pt}
\STATE Select $M$ samples from $\{X^i_k\}$ as initial cluster centers:\\
\vspace{3pt}
$\{\widehat{X}_1^{(0)},\widehat{X}_2^{(0)},...,\widehat{X}_M^{(0)}\}$
\vspace{3pt}
\STATE Set $q=0$ \\
\vspace{3pt}
\STATE \textbf{while} $q<max\_iter$ \textbf{do}\\
\vspace{3pt}
\STATE ~~~~Assign each $X^i_k$ to the nearest cluster center by:\\
\vspace{3pt}
~~~~$d_g^2(X^i_k,X_m^{(q)})=||log_{\widehat{X}_m^{(q)}}(X^i_k)||^2$
\vspace{3pt}
\STATE ~~~~Update the cluster centers using Algorithm~\ref{alg:algKarcher} as\\
\vspace{3pt}
~~~~$\{\widehat{X}_1^{(q+1)},\widehat{X}_2^{(q+1)},...,\widehat{X}_M^{(q+1)}\}$
\vspace{3pt}
\STATE ~~~~$q=q+1$
\vspace{3pt}
\STATE \textbf{end while}
\end{algorithmic}
\end{algorithm}

\textbf{Riemannian GMM}. It has been studied how Gaussian distribution and mixture model can be extended to Riemannian manifold intrinsically in an early work \cite{pennec2006intrinsic}. However the method in this work is computationally expensive due to the nested iterations. Here we employ the similar idea to \cite{faraki2014fisher}, that is, embedding the mid-level words from Riemannian manifold to vector space via explicit mapping function. Thus the training of GMM can be much faster by employing Euclidean techniques.

Generally, through a mapping $\Phi$ from Riemannian manifold to Euclidean space, the Riemannian GMM can be represented as:
\begin{equation}
\label{equ:equGMM}
P(\Phi(X^i_k)|\lambda)=\displaystyle{\sum_{m=1}^M}w_k G(\Phi(X^i_k)|\mu^{\Phi}_m,\Sigma^{\Phi}_m).
\end{equation}

Specifically, for linear subspace $U^i_k$, the mapping from Grassmann manifold to Euclidean space can be represented through the projection metric as indicated in \cite{hamm2008grassmann} as
\begin{equation}
\Phi:\Phi_U=vec(U^i_k {U^i_k}^T).
\end{equation}


And for SPD matrices $C^i_k$ (and $G^i_k$), the mapping to vector space is equivalent to embedding the manifold into its tangent space at identity matrix \cite{arsigny2007geometric}, i.e.
\begin{equation}
\Phi:\Phi_C=vec(log(C^i_k)).
\end{equation}

\subsection{Encoding on Riemannian manifold}

\textbf{BoVW}. For Bag of Visual Words (BoVW) encoding, we compute the geodesic distance between $M$ codewords $\{\widehat{X}_1,\widehat{X}_2,...,\widehat{X}_M\}$ obtained by K-Karcher-means and all $K$ mid-level words $X^i_k (k=1,2,...,K)$ from each video sample $V^i$. The final signature is obtained by soft-assignment. Specifically, as the distances represent the length of curves along the Riemannian manifold which can not be summed up directly, we normalize the distances for each codeword and retain all $M\times K$ values for final representation.

\textbf{VLAD}. Vector of Locally Aggregated Descriptors (VLAD) is to accumulate the difference of the vectors assigned to each codeword, which characterizes the distribution of the vectors with respect to the center \cite{jegou2010aggregating}. Thus, for VLAD, we need to vectorize the codewords and sample mid-level words using mappings $\Phi$. Assuming the vectorized words to be $D$-dimensional, the dimension of our representation is $M \times D$. The accumulated vector for the $m$-th codeword, denoted as $a_m$, can be computed by
\begin{equation}
a_m = \sum_{NN(X^i_k)=\widehat{X}_m}(\Phi(X^i_k)-\Phi(\widehat{X}_m)).
\end{equation}
After concatenating $\{a_1,...,a_M\}$, the whole vector is subsequently $L_2$-normalized according to the original setting in \cite{jegou2010aggregating}.

\textbf{Fisher Vector}. With the Riemannian GMM learned via Equation~\ref{equ:equGMM}, the Fisher score (Fisher Vector is obtained by concatenating the Fisher scores) \cite{jaakkola1999exploiting} for the $m$-th component is computed as
\begin{equation}
\mathcal{G}_m=\frac{1}{K}\sum_{k=1}^K \nabla_\lambda \log G(\Phi(X^i_k)|\mu^{\Phi}_m,\Sigma^{\Phi}_m)
\end{equation}

Specifically, assuming that the covariance matries $\Sigma^{\Phi}_m$ are diagonal, the Fisher encoding can be derived as:
\begin{equation}
\mathcal{G}_{\mu^{\Phi},m}=\frac{1}{K\sqrt{w_m}}\sum_{k=1}^K \gamma_k(m) (\frac{\Phi(X^i_k)-\mu^{\Phi}_m}{\sigma^{\Phi}_m}),
\end{equation}
\begin{equation}
\mathcal{G}_{\sigma^{\Phi},m}=\frac{1}{K\sqrt{2w_m}}\sum_{k=1}^K \gamma_k(m) (\frac{(\Phi(X^i_k)-\mu^{\Phi}_m)^2}{(\sigma^{\Phi}_m)^2}).
\end{equation}
where $\gamma_k(m)$ is the soft assignment to the $m$-th component:
\begin{equation}
\gamma_k(m)=\frac{w_mG(\Phi(X^i_k)|\mu^{\Phi}_m,\Sigma^{\Phi}_m)}{\sum_{m=1}^M w_mG(\Phi(X^i_k)|\mu^{\Phi}_m,\Sigma^{\Phi}_m)}.
\end{equation}

After calculating the Fisher Vector, $L_2$-normalization and power normalization are performed as in \cite{perronnin2010improving} to generate the whole video representation. In the final stage, linear SVM classifier is employed for action recognition.

\section{Experiments}

Our method is evaluated in two tasks on four popular realistic datasets: action recognition task on YouTube, UCF50, and HMDB51 databases; action similarity labeling task on ASLAN database.

\subsection{Datasets}

\textbf{YouTube Action Database} \cite{liu2009recognizing} is collected from YouTube web videos. It contains 11 action categories and a total of 1,168 video sequences of 25 pre-defined groups. Following the general setup used in \cite{liu2009recognizing,le2011learning,wang2013dense}, we adopt Leave-One-Group-Out (LOGO) cross-validation and report the average accuracy over all categories.

\textbf{UCF50 Database} \cite{reddy2013recognizing} is an extension of the YouTube Action Database, which contains totally 6,618 video clips taken from the YouTube website. There are 50 action categories ranging from general sports to daily life exercises. For each category, the video samples are split into 25 groups. We apply the same LOGO protocol as for the YouTube dataset and report average accuracy over all categories.

\textbf{HMDB51 Database} \cite{kuehne2011hmdb} is a large dataset collected from a variety of sources ranging from movies to web videos. It contains a total of 6,766 video sequences and 51 action categories depicting different level of actions, including simple facial actions, general body movements and human interactions. We follow the protocol adopted in \cite{kuehne2011hmdb,wang2013action}, and conduct recognition on three pre-defined train-test splits respectively. The mean avarage accuracy over the three splits is reported for performance measure.
\begin{figure*}[tbh]
\centering
\includegraphics[height=4.5cm]{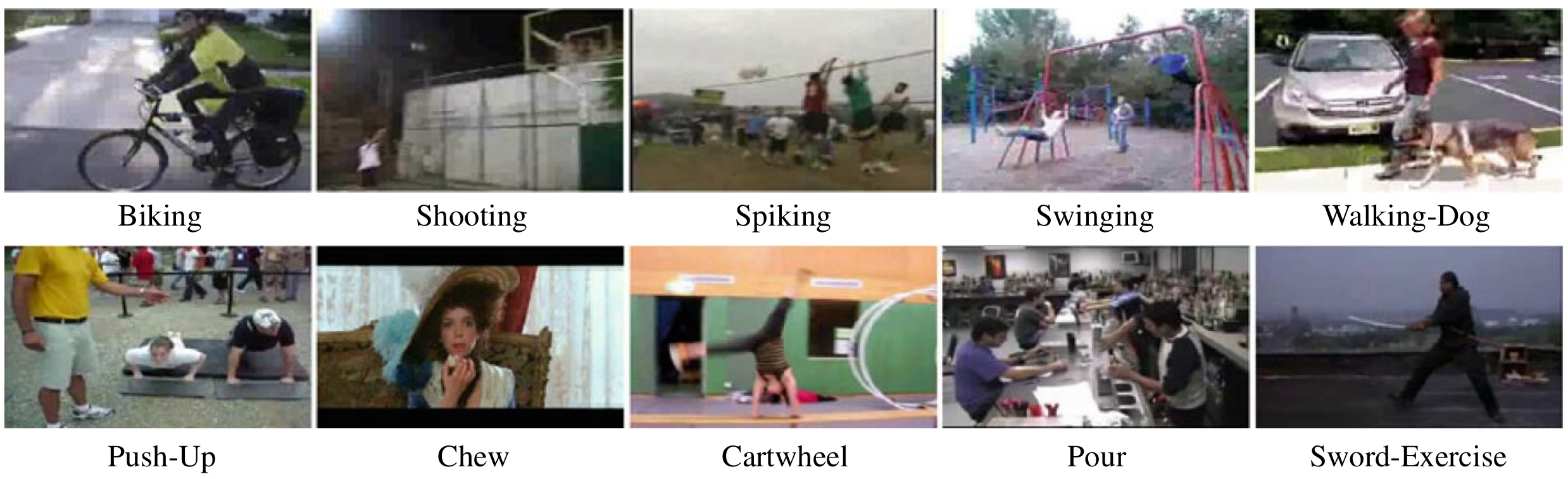}
\caption{Examples from YouTube (upper) and HMDB51 (lower) databases.}
\label{figure:figYHexamples}
\end{figure*}

\begin{figure}[tbh]
\centering
\includegraphics[height=6.5cm]{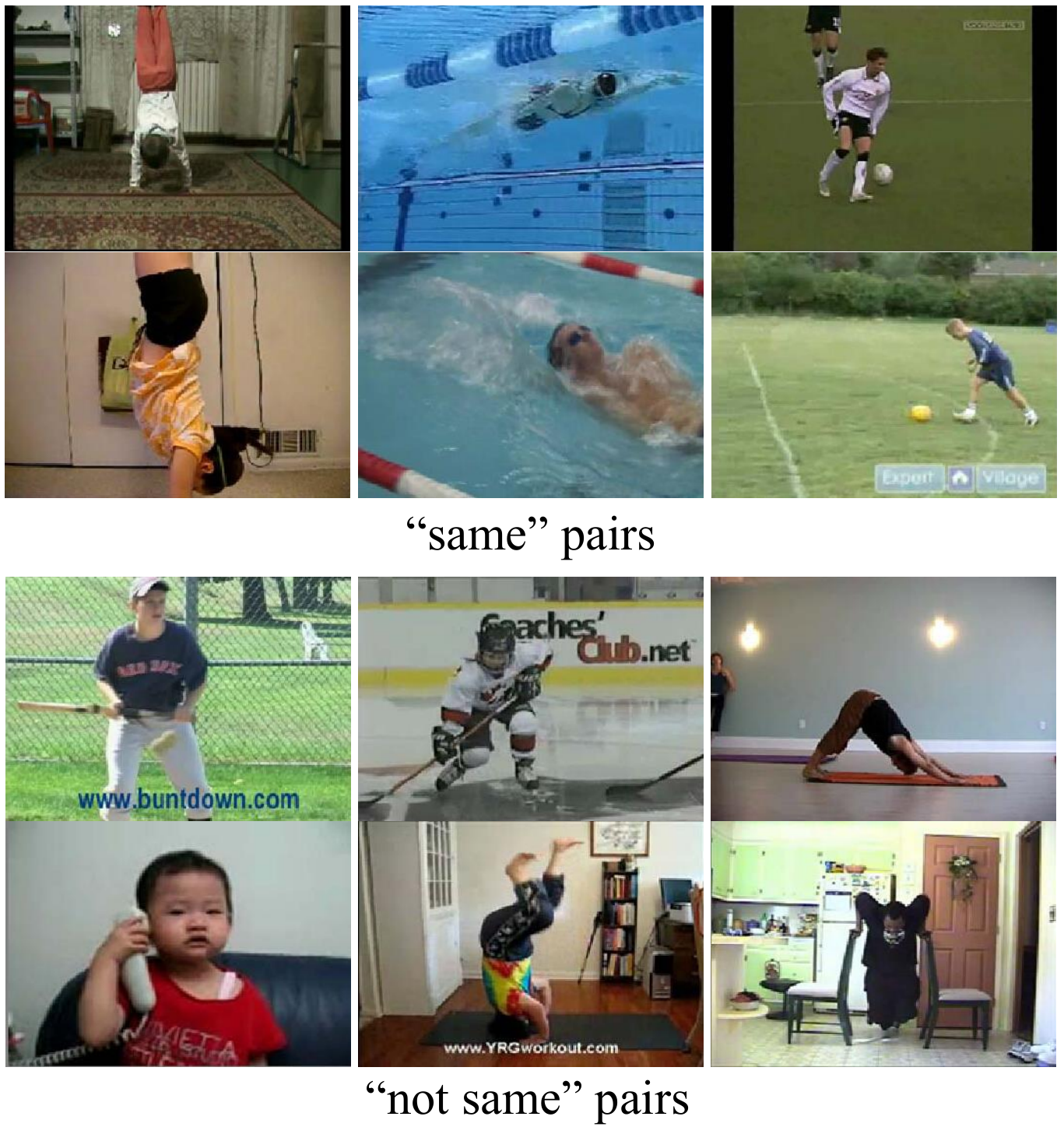}
\caption{Examples of labeled pairs from ASLAN database.}
\label{figure:figAexamples}
\end{figure}
\textbf{ASLAN Database} \cite{kliper2012action} contains 3,631 action videos collected from the web, in a total of 432 categories. The benchmarks for ASLAN are organized into two ``Views''. View-1 is for validation and View-2 is for reporting performance. In this paper, we use the protocol of View-2, which is the 10-fold cross-validation. Specifically, the database is divided into 10 subsets, each of which contains 600 action video pairs: 300 ``same'' and 300 ``not same''. In each fold, 1 subset is for testing and the rest 9 subsets are for training. The Area Under ROC Curve (AUC) and average accuracy over 10-fold are reported for performance comparison.

\subsection{Parameter settings}
For dense features extraction, we use the original scale and the sampling step is 8 pixels in both space and time. The size of space-time grid is 32x32x15, where 32 is the spatial pixels and 15 is the number of frames. The grid is divided into 2x2 cells spatially and 3 cells temporally, which results in 2x2x3 cells totally. The descriptors (i.e. HOG, HOF, MBH) are computed in each cell respectively. The number of histogram bins are set to be 8 for HOG and MBH (MBH is computed separately for horizontal and vertical components, which results two descriptors MBHx, MBHy), and 9 for HOF (including zero bin). The final feature dimensions of the whole grid is 2x2x3x8=96 for HOG, MBHx, MBHy, and 2x2x3x9=108 for HOF.
\begin{table*}[tbh]
\linespread{1.2}
\small
\caption{Comparison of different descriptors and mid-level words modeling methods on YouTube, UCF50, and HMDB51 databases.}
\centering
\begin{tabular}{c|c|ccc|ccc|ccc}
  \hline
  \multicolumn{2}{c|}{\multirow{2}{*}{Datasets \& Methods}} & \multicolumn{3}{c|}{~~~YouTube} & \multicolumn{3}{c|}{~~~UCF50} & \multicolumn{3}{c}{~~~HMDB51} \\
  \cline{3-11}
  \multicolumn{2}{c|}{}          & BoVW & VLAD &~~FV~~& BoVW & VLAD &~~FV~~& BoVW & VLAD &~~FV~~\\
  \hline\hline
  \multirow{5}{*}{HOG} & Origin  & 65.2 & 71.2 & 76.8 & 48.3 & 67.6 & 75.8 & 17.6 & 22.5 & 27.8 \\
                       & SUB     & 64.5 & 75.5 & 79.1 & 53.4 & 68.1 & 73.5 & 22.3 & 31.3 & 32.2 \\
                       & COV     & 64.0 & 79.5 & 81.1 & 54.9 & 72.0 & 76.5 & 23.1 & 31.8 & 33.7 \\
                       & GAU     & 66.0 & 77.9 & 80.2 & 54.2 & 72.1 & 75.6 & 23.8 & 32.9 & 35.3 \\
                       & Combined& 74.4 & 82.6 & 84.3 & 65.2 & 77.4 & 81.0 & 29.3 & 37.1 & 39.8 \\
  \hline
  \multirow{5}{*}{HOF} & Origin  & 60.6 & 69.2 & 73.5 & 46.8 & 66.9 & 73.6 & 24.4 & 33.1 & 37.1 \\
                       & SUB     & 64.4 & 72.8 & 76.9 & 64.6 & 72.8 & 75.4 & 29.5 & 39.7 & 41.1 \\
                       & COV     & 61.2 & 77.7 & 79.2 & 56.5 & 74.4 & 77.0 & 28.2 & 41.2 & 43.0 \\
                       & GAU     & 61.9 & 76.6 & 78.1 & 56.7 & 73.6 & 76.6 & 27.4 & 41.4 & 43.1 \\
                       & Combined& 69.1 & 80.3 & 81.6 & 68.1 & 76.7 & 80.7 & 34.1 & 46.3 & 47.8 \\
  \hline
  \multirow{5}{*}{MBHx}& Origin  & 69.7 & 72.1 & 73.9 & 58.0 & 72.9 & 75.6 & 24.9 & 27.0 & 34.6 \\
                       & SUB     & 58.5 & 68.1 & 70.7 & 44.2 & 67.2 & 67.5 & 16.5 & 24.1 & 24.6 \\
                       & COV     & 57.7 & 73.2 & 74.7 & 53.2 & 73.2 & 73.2 & 18.1 & 28.5 & 32.0 \\
                       & GAU     & 57.9 & 72.9 & 74.9 & 52.1 & 75.0 & 74.9 & 18.6 & 27.3 & 32.7 \\
                       & Combined& 72.1 & 77.9 & 80.9 & 65.2 & 75.8 & 80.6 & 28.7 & 33.7 & 40.3 \\
  \hline
  \multirow{5}{*}{MBHy}& Origin  & 69.1 & 75.3 & 75.2 & 66.0 & 77.5 & 79.1 & 28.5 & 32.7 & 39.3 \\
                       & SUB     & 54.5 & 60.8 & 65.5 & 45.0 & 67.5 & 67.3 & 16.9 & 23.4 & 25.1 \\
                       & COV     & 52.0 & 68.1 & 72.9 & 55.8 & 72.6 & 72.2 & 21.4 & 28.4 & 31.9 \\
                       & GAU     & 53.1 & 70.0 & 73.7 & 54.2 & 74.7 & 77.7 & 23.8 & 29.2 & 33.5 \\
                       & Combined& 70.5 & 79.9 & 81.3 & 70.6 & 79.7 & 83.9 & 32.1 & 37.4 & 44.6 \\
  \hline
  \multicolumn{2}{c|}{\textbf{Late Fusion}}& 83.1 & 89.2 & \textbf{90.3} & 83.9 & 88.2 & \textbf{90.7} & 45.3 & 53.6 & \textbf{56.4} \\
  \hline
\end{tabular}
\label{table:tabYUHresults}
\end{table*}

\begin{figure*}
\centering
\subfigure[VLAD (HOG)]{
\includegraphics[height=3.2cm]{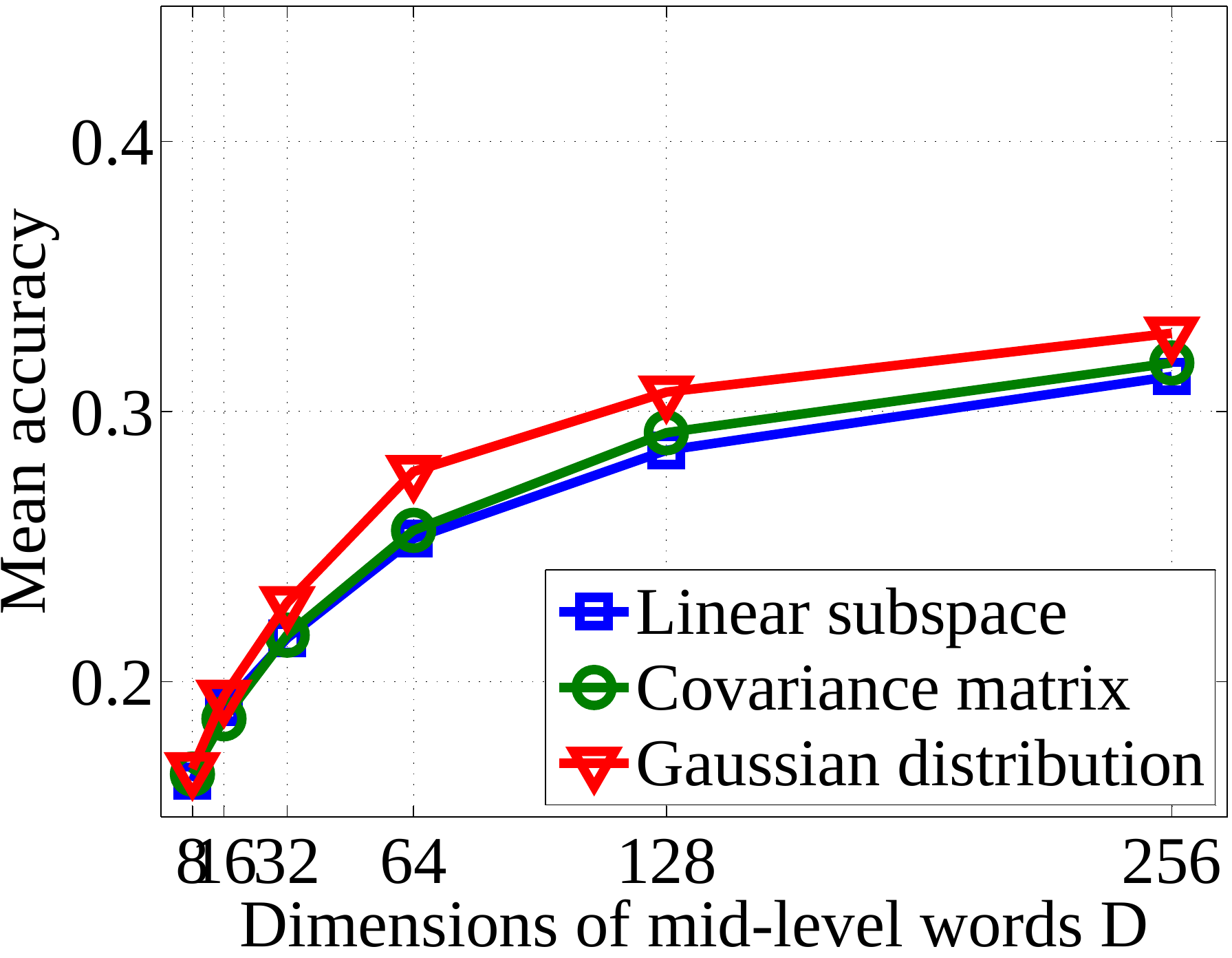}}
\subfigure[FV (HOG)]{
\includegraphics[height=3.2cm]{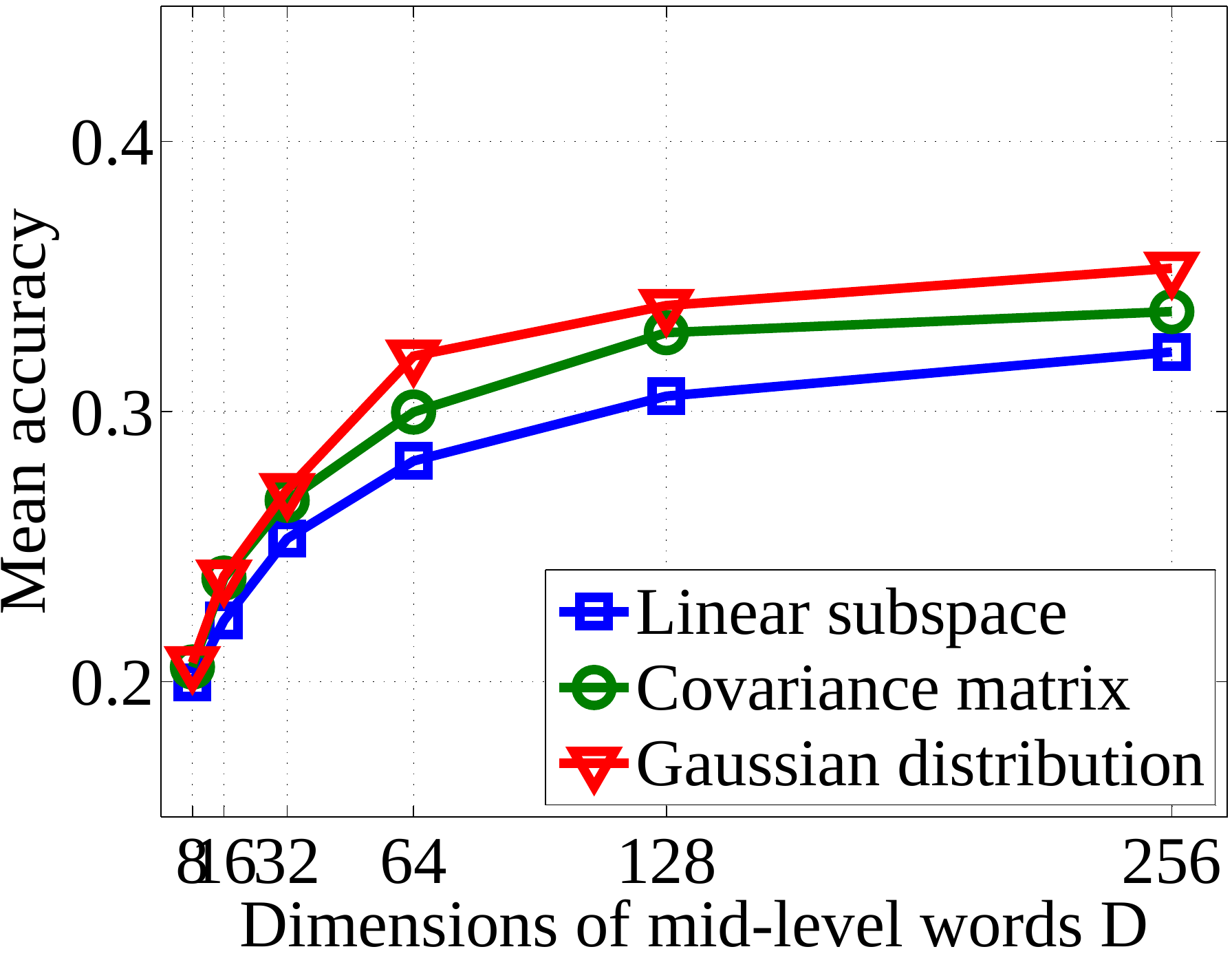}}
\subfigure[VLAD (HOF)]{
\includegraphics[height=3.2cm]{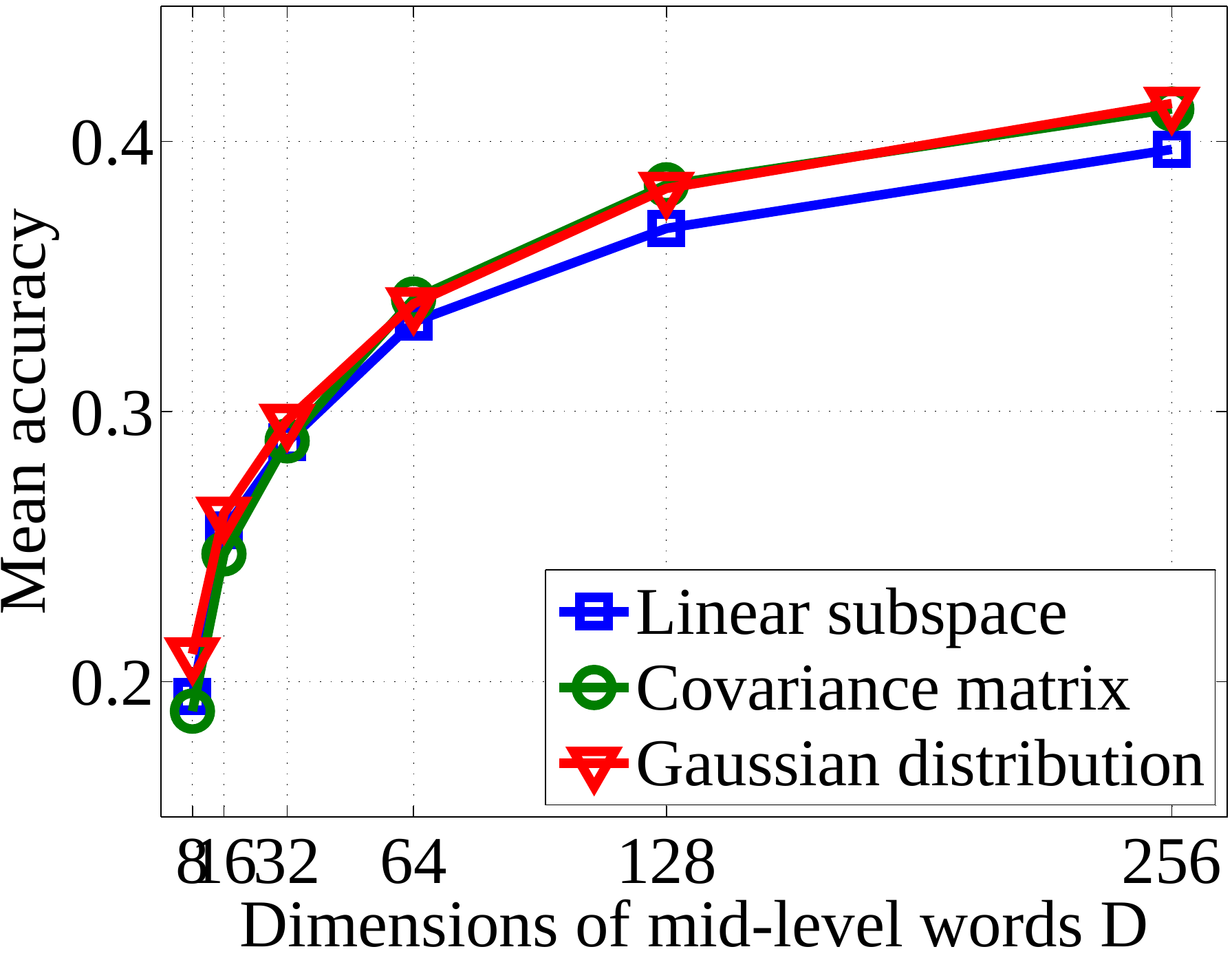}}
\subfigure[FV (HOF)]{
\includegraphics[height=3.2cm]{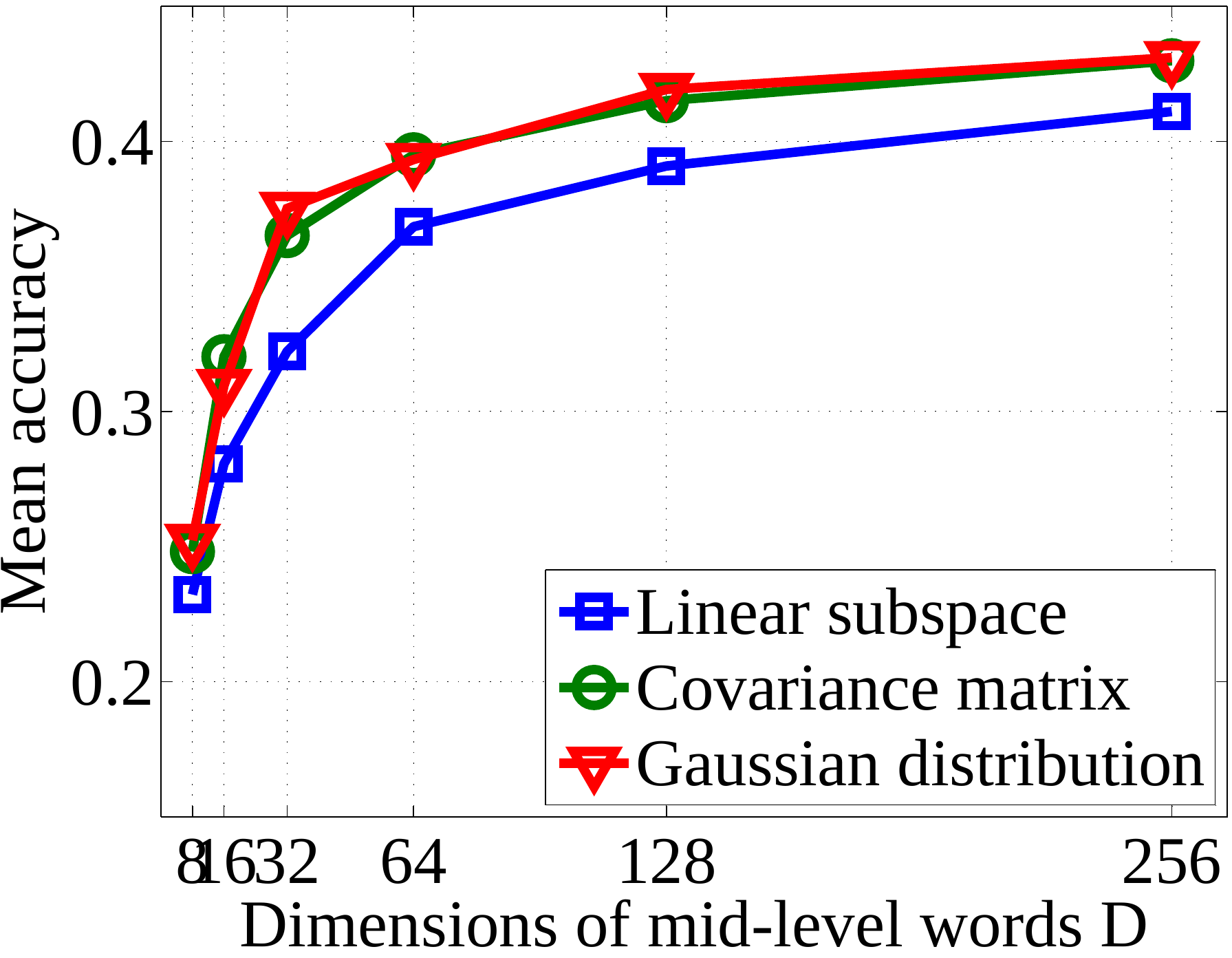}}
\caption{Recognition accuracy with different dimensions of mid-level words $D$ on HMDB51 database. (a) VLAD (HOG) (b) FV (HOG) (a) VLAD (HOF) (b) FV (HOF). \textbf{Best viewed in color}.}
\label{fig:figParamD}
\end{figure*}

\begin{figure*}[tbh]
\centering
\subfigure[VLAD (HOG)]{
\includegraphics[height=3.2cm]{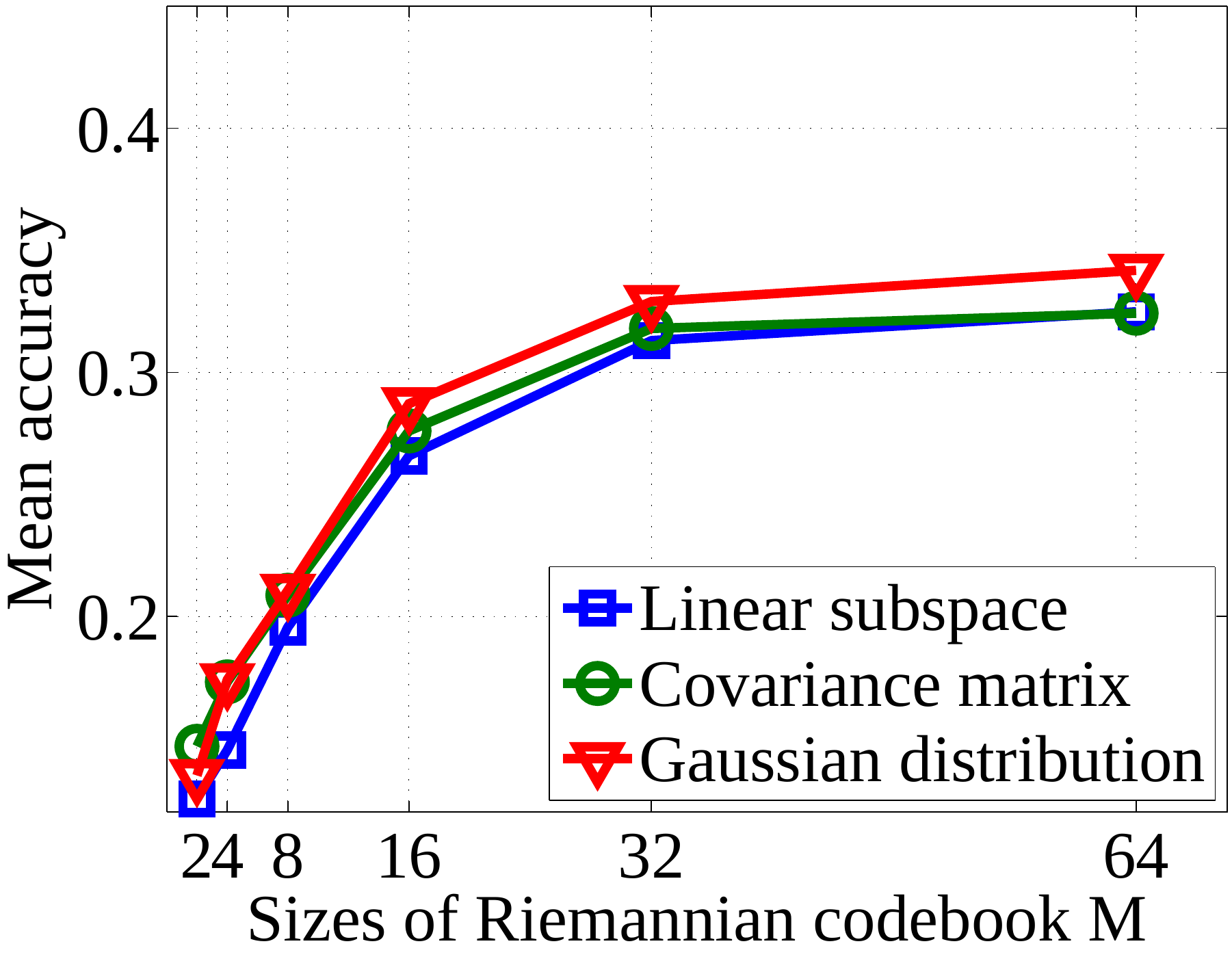}}
\subfigure[FV (HOG)]{
\includegraphics[height=3.2cm]{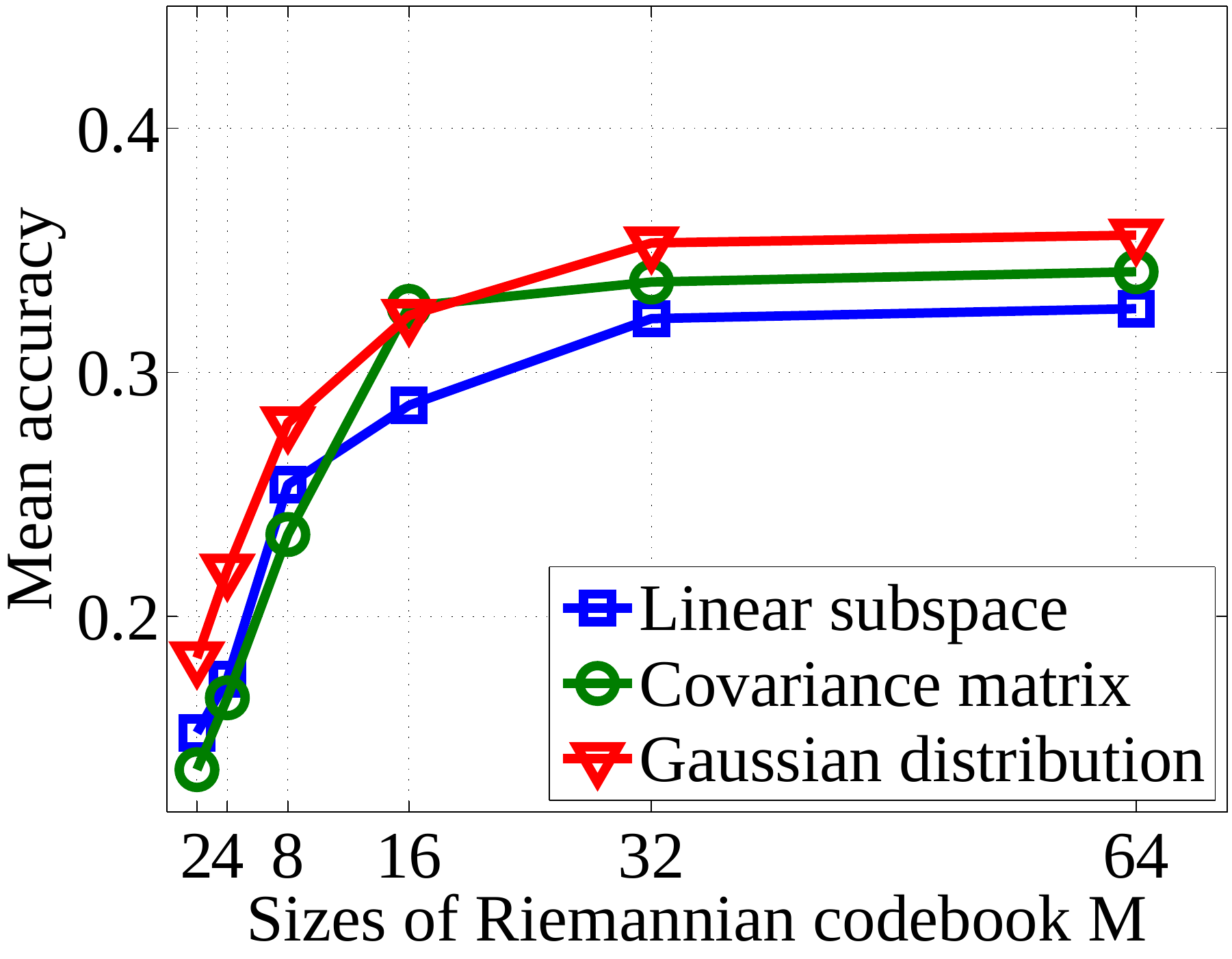}}
\subfigure[VLAD (HOF)]{
\includegraphics[height=3.2cm]{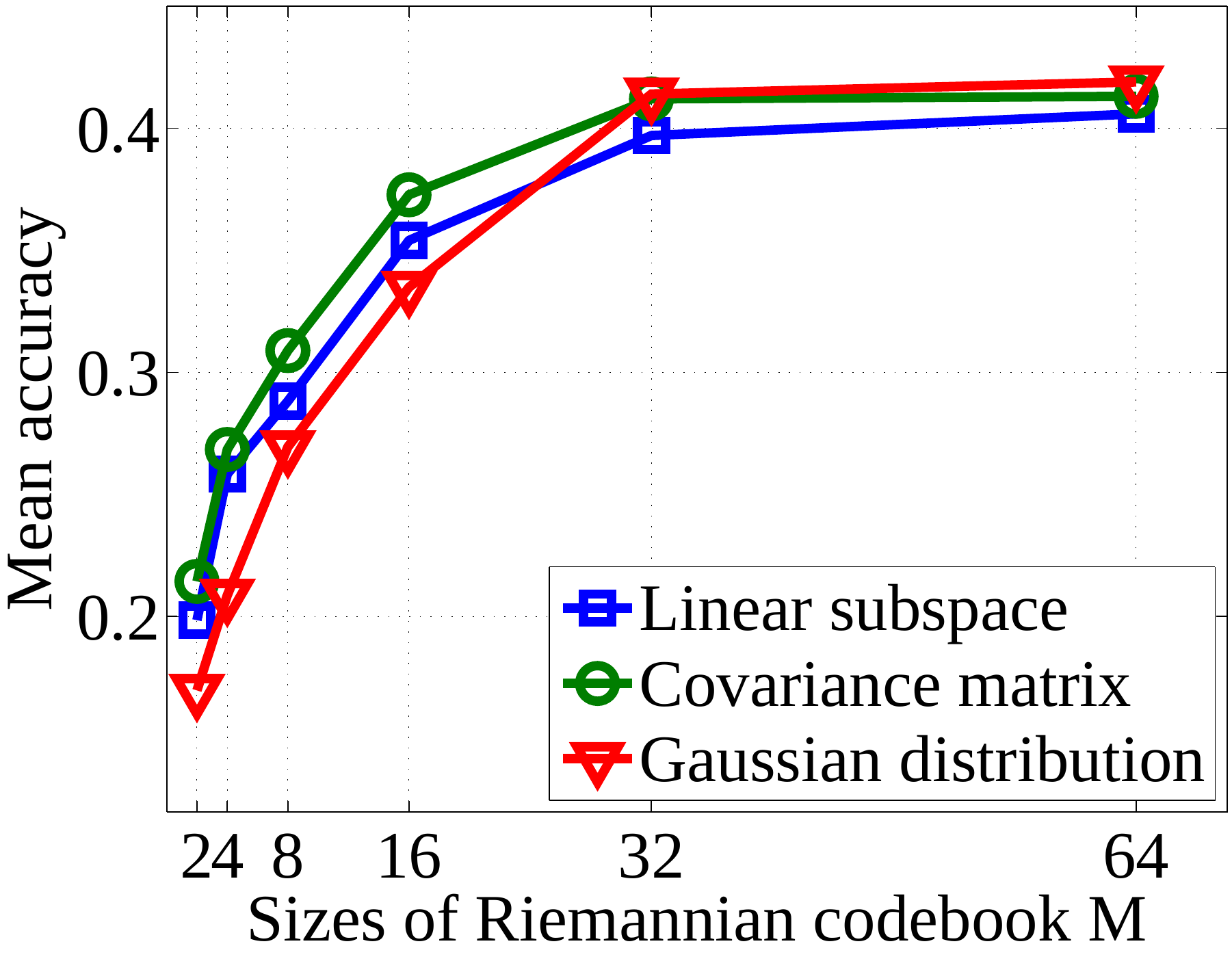}}
\subfigure[FV (HOF)]{
\includegraphics[height=3.2cm]{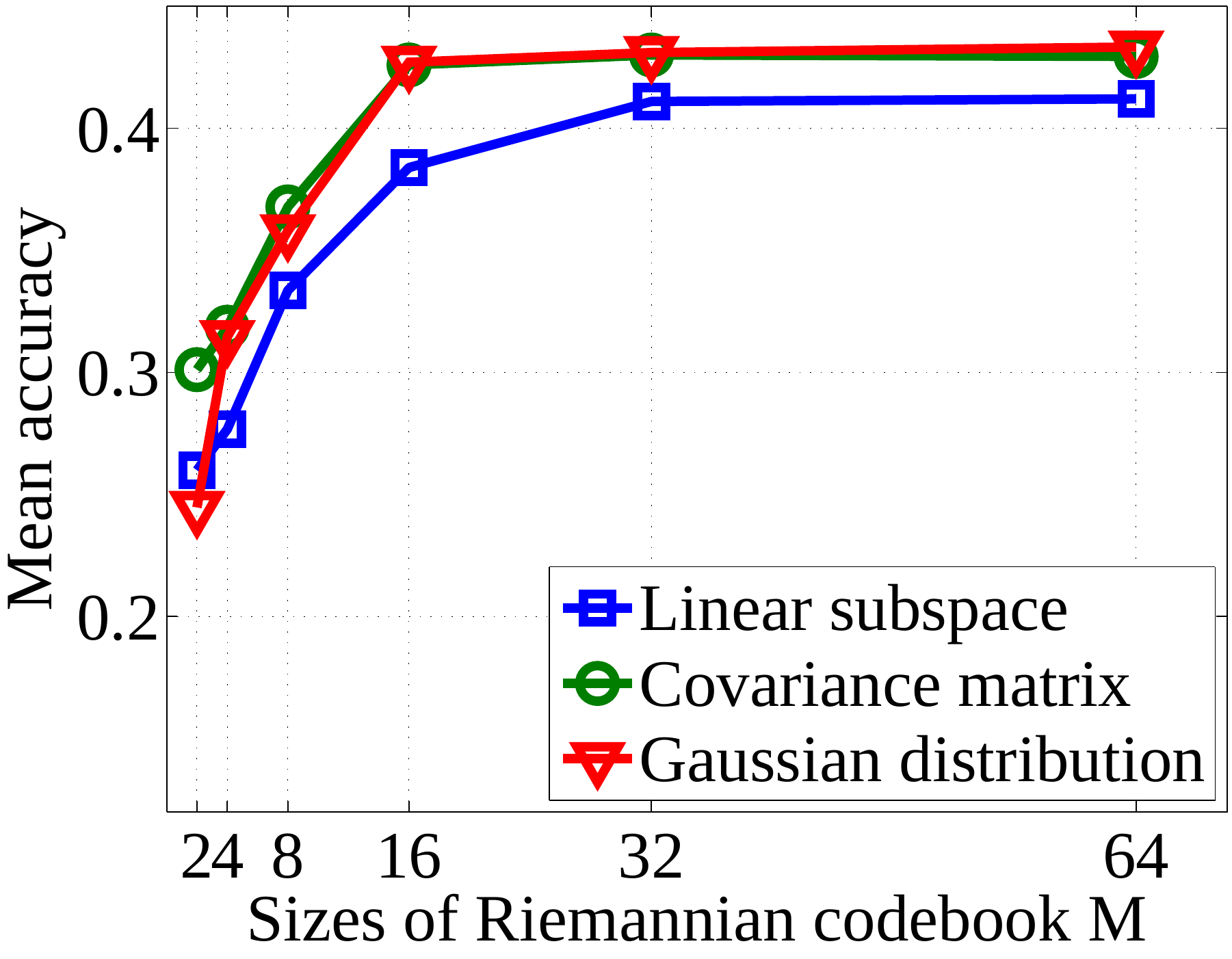}}
\caption{Recognition accuracy with different sizes of Riemannian codebooks $M$ on HMDB51 database. (a) VLAD (HOG) (b) FV (HOG) (a) VLAD (HOF) (b) FV (HOF). \textbf{Best viewed in color}. }
\label{fig:figParamM}
\vspace{-5pt}
\end{figure*}

To learn the universal GMM, we first employ PCA to reduce the feature dimensions by a factor of 0.5 (i.e. HOG/MBHx/MBHy from 96 to 48, HOF from 108 to 54), then set the number of Gaussian components $K=256$ and the number of features assigned to each component $T=64$. Three different statistics are employed on each feature group for mid-level words modeling. Two important parameters in the following steps are studied: (1) The dimensions of the mid-level words $D$: In VLAD and FV encoding, the mid-level words are mapped to vector space via $\Phi$ and the dimensions are reduced to $D$ (see Section 3.2) via PCA. (2) The size of Riemannian codebook $M$: i.e. the number of K-Karcher-means clusters in Algorithm~\ref{alg:algK} and the number of Gaussian components in Equation~\ref{equ:equGMM}. The relations between recognition performance and each of the two parameters are presented in Figure~\ref{fig:figParamD} and Figure~\ref{fig:figParamM} respectively. As for validation purpose only, we conduct such experiments on HMDB51 database using VLAD and FV based on HOG and HOF features.

In all figures, the blue, green, and red curves represent the mid-level words modeling schemes via linear subspace, covariance matrix, and Gaussian distribution respectively. We can observe the rising trend of all the curves as $D$ and $M$ increase. For computational efficiency, we choose $D=256$ and $M=32$ for both VLAD and FV on all types of descriptors. And for BoVW, we use the same $D$ while setting $M$ as 64. In classification stage, we use a linear SVM with the parameter $C=100$.

\subsection{Results comparison}

In the experiments, we evaluate our method on two tasks: action recognition and action similarity labeling. For recognition, the video representations are directly fed to SVM for classification. While for similarity labeling, we compute the distances between each video sample pair and concatenate the values into feature vectors in the same way as in \cite{kliper2012action}. We demonstrate the action recognition performance on YouTube, UCF50, and HMDB51 databases in Table~\ref{table:tabYUHresults}. For each type of low-level features, we conduct experiments based on both the original descriptor and different modelings of mid-level words. Here the codebook sizes for BoVW, VLAD, and FV on low-level descriptors are 1024, 256, 256 respectively. ``SUB'', ``COV'', and ``GAU'' are short for linear subspace, covariance matrix, and Gaussian distribution. We can observe that for HOG and HOF, the mid-level words can consistently bring significant improvement under different encoding methods on both databases. However for MBHx/MBHy, the performance shows degradation when introducing the mid-level words (especially for BoVW encoding). We mildly conjecture that it is caused by the fact that MBH encodes the \textit{\textbf{relative motions}} occurring in the boundary of foreground and background, which refer to different basis and make it inappropriate to be statistically measured in a unified feature space (while HOG/HOF features characterize the \textit{\textbf{absolute variations}} of appearance or motion occurring in the unified space), thus deteriorating the statistic models for mid-level words construction. As a whole, the combination of low-level descriptor and mid-level words results in good performance for their description complementarity.

Table~\ref{tab:tabAresults} demonstrates the action similarity labeling results on ASLAN database. It can be observed that the improvement brought by mid-level words is not as significant as that on recognition task. The reason may be that the proposed mid-level word based representations are not very readily used for distance computation in Euclidean space for decision without considering the inherent Riemannian manifold geometric information. This also motivates us to explore more tailored Riemannian manifold metric learning method in such similarity labeling task for further improvement.
\begin{table*}[tbh]
\linespread{1.2}
\small
\caption{Comparison of different descriptors and mid-level words modeling methods on ASLAN database.}
\centering
\begin{tabular}{c|c|cc|cc}
\hline
\multicolumn{2}{c|}{\multirow{2}{*}{Method}} & \multicolumn{2}{c|}{VLAD} & \multicolumn{2}{c}{FV} \\
\cline{3-6}
\multicolumn{2}{c|}{}         & Accuracy (\%) & AUC (\%) & Accuracy (\%) & AUC (\%) \\
\hline\hline
\multirow{2}{*}{HOG} & Origin & 59.43$\pm$1.86 & 63.02 & 60.17$\pm$1.74 & 63.80 \\
                     & Combined \scriptsize{(SUB+COV+GAU)} & 61.28$\pm$1.87 & 66.21 & 62.23$\pm$1.79 & 67.20 \\
\hline
\multirow{2}{*}{HOF} & Origin & 59.20$\pm$1.74 & 62.90 & 59.70$\pm$1.68 & 63.15 \\
                     & Combined \scriptsize{(SUB+COV+GAU)} & 61.63$\pm$1.54 & 66.32 & 62.10$\pm$1.60 & 66.58 \\
\hline
\multirow{2}{*}{MBH} & Origin & 58.65$\pm$1.82 & 61.92 & 59.32$\pm$1.63 & 63.63 \\
                     & Combined \scriptsize{(SUB+COV+GAU)} & 60.83$\pm$1.87 & 65.30 & 61.73$\pm$1.66 & 66.76 \\
\hline
\multicolumn{2}{c|}{\textbf{Late Fusion}} & \textbf{63.32$\pm$1.79} & \textbf{69.68} & \textbf{65.17$\pm$1.93} & \textbf{70.29} \\
\hline
\end{tabular}
\label{tab:tabAresults}
\end{table*}
In the end, we compare our method with the state-of-the-art on the three databases in Table~\ref{tab:tabState}. All of the comparison results are directly cited from the original literatures. The second group shows the experimental results of some mid-level representation based methods, i.e. \cite{sapienza2014learning}, \cite{wang2013motionlets}, \cite{zhu2013action}, which are just our competitor. Note that, in Table~\ref{tab:tabASLAN} on similarity labeling task, for fair comparison, we only compare our method with the unsupervised scheme without metric learning or discriminative learning when computing sample pair distances. In all cases, our results achieve very competitive performance with these most recent works.
\begin{table*}[tbh]
\small
\caption{Comparison with the state-of-the-art results on three databases.}
\centering
\subtable[YouTube]{
\begin{tabular}{c|c}
\hline
Method & Accuracy(\%) \\
\hline\hline
\textit{Liu et al. }\cite{liu2009recognizing} & 71.2 \\
\textit{Brendel et al.} \cite{brendel2010activities} & 77.8 \\
\textit{Wang et al.} \cite{wang2013dense} & 85.4 \\
\textit{Yang et al.} \cite{yang2014action} & 88.0 \\
\hline
Action-parts \cite{sapienza2014learning} & 84.5 \\
Action-Gons \cite{wang2014action} & 89.7 \\
Actons \cite{zhu2013action} & 89.4 \\

\hline
\textbf{Our method} & 90.3 \\
\hline
\end{tabular}
}
\subtable[UCF50]{
\begin{tabular}{l|c}
\hline
Method & Accuracy(\%) \\
\hline\hline
\textit{Kliper-Gross et al.} \cite{kliper2012motion} & 72.7 \\
\textit{Reddy et al.} \cite{reddy2013recognizing} & 76.9 \\
\textit{Shi et al.} \cite{shi2013sampling} & 83.3 \\
\textit{Wang et al.} \cite{wang2013dense} & 85.6 \\
\textit{Wang et al.} \cite{wang2013action} & 91.2 \\
\hline
Motionlets \cite{wang2013motionlets} & 73.9 \\
Motion-atoms \cite{wang2013mining} & 85.7 \\
\hline
\textbf{Our method} & 90.7 \\
\hline
\end{tabular}
}
\subtable[HMDB51]{
\begin{tabular}{c|c}
\hline
Method & Accuracy(\%) \\
\hline\hline
\textit{Kliper-Gross et al.} \cite{kliper2012motion} & 29.2 \\
\textit{Wang et al.} \cite{wang2013dense} & 46.6 \\
\textit{Jian et al.} \cite{jain2013better} & 52.1 \\
\textit{Hou et al.} \cite{hou2014damn} & 57.9 \\
\hline
Action-parts \cite{sapienza2014learning} & 37.2 \\
Motionlets \cite{wang2013motionlets} & 42.1 \\
Actons \cite{zhu2013action} & 54.0 \\
\hline
\textbf{Our method} & 56.4 \\
\hline
\end{tabular}
}
\label{tab:tabState}
\end{table*}

\begin{table}[tbh]
\small
\caption{Comparison with the state-of-the-art results on ASLAN.}
\centering
\begin{tabular}{l|cc}
\hline
Method & Accuracy(\%) & AUC (\%)\\
\hline\hline
\textit{Kliper-Gross et al.} \cite{kliper2012action} & 60.88 & 65.30 \\
\textit{Kliper-Gross et al.} \cite{kliper2012motion} & 64.27 & 69.20 \\
\textit{Peng et al.} \cite{peng2014large} \scriptsize{(VLAD)} & 61.38 & 66.39 \\
\textit{Peng et al.} \cite{peng2014large} \scriptsize{(FV)} & 63.75 & 69.28 \\
\hline
\textbf{Our method} & \textbf{65.17} & \textbf{70.29} \\
\hline
\end{tabular}
\label{tab:tabASLAN}
\end{table}

\subsection{Computation time}

To evaluate computation time, we report the average for each video sample using one PC with 3.40GHz CPU and 16G RAM. The low-level descriptors extraction takes 11.1s. For each of the 4 descriptors, our two-stage encoding takes 8.2s (including read-write time) and overall takes 43.9s. Compared with other mid-level based methods: Motionlets \cite{wang2013motionlets}: 70s; Action-parts \cite{sapienza2014learning}: 227s; Action bank \cite{sadanand2012action}: 1156s, our method is much faster with higher accuracy.

\section{Conclusions}
In this paper, we propose a novel representation approach by constructing mid-level words in videos and encoding them on Riemannian manifold. Specifically, we first conduct a global alignment on the low-level features to generate a bank of corresponding mid-level words, which are statistically represented as points residing on Riemannian manifolds. Then we innovatively construct intrinsic Riemannian codebooks for encoding of BoVW/VLAD/FV to obtain the mid-level words based video representations. Our method is evaluated in two tasks on four popular realistic datasets and has achieved the state-of-the-art performance in all cases. For future work, we are trying to extend this study in two aspects: (1) Figuring out what characteristics in low-level features, when combined with statistical modeling, can benefit the representation of dynamic variations; (2) Deriving Riemannian-based metric learning for mid-level words based representation. Moreover, as mid-level words can be regarded as sub-units of the actions, our method can be naturally extended to a different range of applications like action localization or spotting.

{\small
\bibliographystyle{ieee}
\bibliography{egbib}
}

\end{document}